\documentclass[journal,twoside,web]{ieeecolor}
\usepackage{generic}
\usepackage{cite}
\usepackage{amsmath,amssymb,amsfonts}
\usepackage{algorithmic}
\usepackage{graphicx}
\usepackage{textcomp}
\usepackage{glossaries}
\usepackage{hyperref}

\setacronymstyle{long-short}
\newacronym{ehr}{EHR}{electronic health record}
\newacronym{rct}{RCT}{randomized clinical trial}
\newacronym{rnn}{RNN}{recurrent neural network}
\newacronym{rnnae}{RNN-AE}{recurrent neural network autoencoder}

\newacronym{fc}{FC}{fully connected}

\newacronym{lstm}{LSTM}{long short-term memory}
\newacronym{gru}{GRU}{gater recurrent neural network}
\newacronym{agru}{aGRU}{adversarial gated recurrent neural network}
\newacronym{tlstm}{tLSTM}{time-aware LSTM}
\newacronym{ae}{AE}{autoencoder}
\newacronym{auc}{AUC}{area under the curve}
\newacronym{mse}{MSE}{mean squared error}
\newacronym{pca}{PCA}{principal component analysis}

\newacronym{ouh}{OUH}{Oxford University Hospital}
\newacronym{chelwest}{ChelWest}{Chelsea and Westminster Hospital}
\newacronym{hf}{HF}{heart failure}

\newacronym{bee}{BeE}{before-event}
\newacronym{s2e}{S2E}{start-to-event}
\newacronym{afe}{AfE}{after-event}
\newacronym{e2e}{E2E}{event-to-end}
\newacronym{all}{ALL}{ALL}
\newacronym{bin}{bin}{binary normalization}
\newacronym{buck}{bucket}{bucketized normalization}
\newacronym{minmax}{minmax}{min-max normalization}
\newacronym{rank}{rank}{rank normalization}

\newacronym{rmse}{RMSE}{root mean squared error}
\newacronym{knn}{k-NN}{K-nearest neighbors}
\newacronym{ofm}{OFM}{objective function mismatch}

\usepackage{booktabs}
\usepackage[load-configurations=version-1]{siunitx} % newer version 
\newcommand{\acks}[1]{\section*{Acknowledgments}#1}

\def\BibTeX{{\rm B\kern-.05em{\sc i\kern-.025em b}\kern-.08em
    T\kern-.1667em\lower.7ex\hbox{E}\kern-.125emX}}
\begin{document}
\title{Compensating trajectory bias for unsupervised patient stratification using adversarial recurrent neural networks}
\author{Avelino Javer\textsuperscript{[1]}, Owen Parsons\textsuperscript{[1]}, Oliver Carr\textsuperscript{[1]}, Janie Baxter\textsuperscript{[1]}, Christian Diedrich\textsuperscript{[2]}, Eren El\c{c}i\textsuperscript{[3]}, Steffen Schaper\textsuperscript{[2]}, Katrin Coboeken\textsuperscript{[2]} and Robert D{\"u}richen\textsuperscript{[1]}. 
% \thanks{Submitted for review on 14/12/21 }
\thanks{[1] Sensyne Health Plc, Oxford, UK}
\thanks{[2] Systems Pharmacology \& Medicine, Bayer AG -  Pharmaceuticals, Wuppertal, Germany}
\thanks{[3] Biomedical Data Science, Bayer AG - Pharmaceuticals, Wuppertal, Germany}}

% \author{
% Avelino Javer\authorrefmark{[1]},
% Owen Parsons\authorrefmark{[1]},
% Oliver Carr\authorrefmark{[1]},
% Janie Baxter\authorrefmark{[1]},
% Christian Diedrich\authorrefmark{[2]},
% Eren El\c{c}i\authorrefmark{[3]},
% Steffen Schaper\authorrefmark{[2]},
% Katrin Coboeken\authorrefmark{[2]} and
% Robert D{\"u}richen\authorrefmark{[1]}
% }

% \thanks{}
% \thanks{[1] Sensyne Health Plc, Oxford, UK}
% \thanks{[2] Systems Pharmacology \& Medicine, Bayer AG -  Pharmaceuticals, Wuppertal, Germany}
% \thanks{[3] Biomedical Data Science, Bayer AG - Pharmaceuticals, Wuppertal, Germany}
% \author{
% \Name{Avelino Javer}\footnotemark[1]{}  \Email{avelino.javer@sensynehealth.com}\\
% \Name{Oliver Carr}\footnotemark[1]{} \Email{oliver.carr@sensynehealth.com}\\
% \Name{Owen Parsons}\footnotemark[1]{}\Email{owen.parsons@sensynehealth.com}\\
% \Name{Janie Baxter}\footnotemark[1]{} \Email{janie.baxter@sensynehealth.com}\\
% \Name{Christian Diedrich}\footnotemark[2]{} \Email{christian.diedrich@bayer.com}\\
% \Name{Eren El\c{c}i}\footnotemark[3]{} \Email{eren.elci@bayer.com}\\
% \Name{Steffen Schaper}\footnotemark[2]{} \Email{steffen.schaper@bayer.com}\\
% \Name{Katrin Coboeken}\footnotemark[2]{} \Email{katrin.coboeken@bayer.com}\\
% \Name{Robert D{\"u}richen}\footnotemark[1]{} \Email{robert.durichen@sensynehealth.com}\\
% }

\maketitle
\begin{abstract}
\Glspl{ehr} are an important source of information which can be used in patient stratification to discover novel disease phenotypes. However, they can be challenging to work with as data is often sparse and irregularly sampled. One approach to solve these limitations is learning dense embeddings that represent individual patient trajectories using a \gls{rnnae}. 
This process can be susceptible to unwanted data biases. 
We show that patient embeddings and clusters using previously proposed \gls{rnnae} models might be impacted by a \textit{trajectory bias}, meaning that 
results are dominated by the amount of data contained in each patients trajectory, instead of clinically relevant details.
We investigate this bias on 2 datasets (from different hospitals) and 2 disease areas as well as using different parts of the patient trajectory. 
Our results using 2 previously published baseline methods indicate a particularly strong bias in case of an \textit{event-to-end} trajectory. 
%In particular, the variation in the amount of data recorded across patients can lead to the embedding models being strongly dominated by the amount of data contained in each patients trajectory, instead of encoding clinical relevant details.
We present a method that can overcome this issue using an adversarial training scheme on top of a \gls{rnnae}. %that result in a reduced \textit{trajectory bias} in the learnt embedding space. 
Our results show that our approach can reduce the \textit{trajectory bias} in all cases. 
\end{abstract}


% \fnsymbol{\footnotetext[1]{Submitted for review on 09/12/21}}
% \fnsymbol{\footnotetext[2]{[1] Sensyne Health Plc, Oxford, UK}}
% \fnsymbol{\footnotetext[3]{[2] Systems Pharmacology \& Medicine, Bayer AG -  Pharmaceuticals, Wuppertal, Germany}}
% \fnsymbol{\footnotetext[4]{[3] Biomedical Data Science, Bayer AG - Pharmaceuticals, Wuppertal, Germany}}

% \thanks{Submitted for review on 01/12/21}
% \thanks{[1] Sensyne Health Plc, Oxford, UK}
% \thanks{[2] Systems Pharmacology \& Medicine, Bayer AG -  Pharmaceuticals, Wuppertal, Germany}
% \thanks{[3] Biomedical Data Science, Bayer AG - Pharmaceuticals, Wuppertal, Germany}

\section{Introduction}
\label{sec:intro}
In recent years, the availability of \gls{ehr} data has increased dramatically. 
Such data contain the medical history of patients across multiple years, with information ranging from primary and secondary diagnosis codes and performed procedures to laboratory values and prescribed medications. 
We refer to the sequence of clinical observations of a patient as patient trajectory. 
Despite being a rich source of information, working with longitudinal real world \gls{ehr} data is challenging. Typically \glspl{ehr} consist of mixed data types (e.g. binary diagnoses codes and continuous laboratory values) and are highly irregularly sampled, sparse, and noisy.

% intro to patient stratification
Alongside clinical objectives, \gls{ehr} data can be used to identify novel patient phenotypes to support the drug development process as well as validation within clinical trials \cite{Evans2021}. 
The aim of patient stratification is to identify clinically meaningful groups of patients which share either similar characteristics along their patient trajectory (e.g. having similar comorbidities or medications prescribed) or clinical outcomes (e.g. mortality or probability of re-hospitalization) \cite{Banda2018}. 
We refer to these two approaches as  \textit{unsupervised} and \textit{outcome based} patient stratification, respectively.

In the case of unsupervised patient stratification, the general approach consists of two main steps: first, learning a dense patient representation, an embedding, of the original patient trajectory, and, second, performing a clustering of patients in the embedded space to identify novel subgroups. Learning a suitable embedding to represent patients medical history is non-trivial due to the longitudinal and noisy nature of \gls{ehr} data, the fact that the data contains mixed data types, and the non linear relationship between the covariates. 

This paper focuses on learning clinically meaningful representations from a large collection of \gls{ehr} by compensating the trajectory bias (defined below) using an \gls{rnnae} approach. 

%The representations can be then used to produce clinically meaningful unsupervised patient clusters.

When using traditional \gls{rnnae} approaches the learned embeddings as well as the identified patient clusters are often strongly determined by how much data is available across the individual trajectory of the patient.
We refer to this as the trajectory bias. 
Even though this information might be an indicator of how healthy a patient is, {\it i.e.} the less data is available, the healthier the patient tends to be in many settings. However, the different amount of available data could also be due to the fact, for example, that the patient relocated and was treated in a different facility. 
Other reasons might be due to different ways of working across wards in a hospital; {\it e.g.} maybe not all data are recorded electronically, or due to the general IT infrastructure; {\it e.g.}, data might get lost when being migrated across platforms.
For these reasons, being able to represent and cluster patient trajectories in ways that are not overly sensitive to variations in the \textit{length} of trajectory is desirable. There is a need to develop  methods where the role of the amount of data within a patient trajectory is reduced and instead shift the focus towards other differentiating features in the data while increasing the robustness against missing information.

\begin{figure}[tb]
\label{fig:sketch_bias} 

  {\caption{Illustration of trajectory bias. In this example, \gls{ehr} data were obtained between 2014-19. Objective is to stratify patients based on data available before a given index event such as the first acute \gls{hf} episode. Due to different amounts of data, patient A and B are more likely to be mapped to different clusters in the embedding space even if they share similar clinical observations. Dotted lines indicate potential result of a clustering algorithm. The colour of the trajectories of patients A, B and C indicates their trajectory length according to the legend at the bottom of the figure.}}
  {\includegraphics[width=1\linewidth]{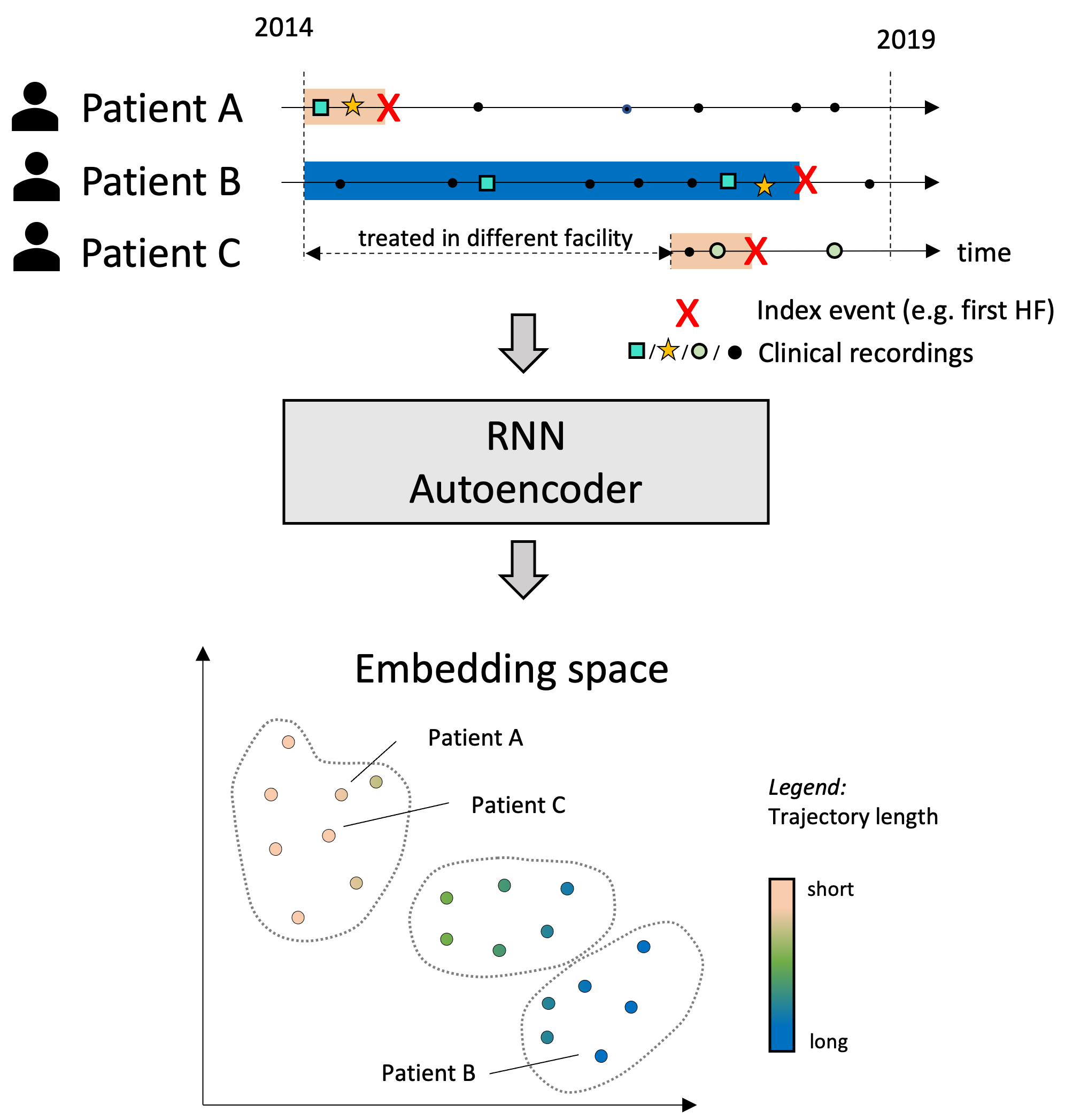}}
 
\end{figure}

Furthermore, the effect of the trajectory bias becomes exacerbated if only specific parts of a patient trajectory are investigated. 
This is illustrated in \hyperref[fig:sketch_bias]{\hyperref[fig:sketch_bias]{Figure 1}}.
Often, \gls{ehr} datasets cover a specific time range, e.g. 2014-2019. 
If, for example, the objective is to stratify \gls{hf} patients based on their trajectory/data before the first acute \gls{hf} event, the trajectory lengths are very different as the event could occur in 2015 as for patient A and in 2018-19 for patient B and C.
Assuming that data for patient B is available from 2014, 
it is less likely that patients A and B would be clustered together due to the different trajectory length even if they have similar clinical observations. 
We present a novel adversarial \gls{rnnae} approach to reduce this trajectory bias within the embedding space, such that resulting clusters are more dependent on medical information rather than on the amount of data present within a time-series.

Our approach is evaluated over its ability to reduce trajectory bias across two cardiovascular \gls{ehr} datasets from different hospitals which includes diagnoses, procedures, medication and laboratory values and a total of $N = 623,364$ patients combined.

\section{Literature}
\label{sec:literature}
% literature review - non temporal embedding of EHR data
In recent years, an increasing amount of machine learning (and in particular deep learning)-based approaches have been investigated to identify novel patient phenotypes \cite{Banda2018}.
The general approach can be grouped into two steps: first, learning a dense patient embedding of the noisy, irregular sampled raw data, and, second, performing a cluster analysis on the embedded data \cite{Baytas2017,Yin2020}. 
%Initial approaches to learn a dense patient embedding were based on the co-occurrence of features using methods such as \textit{word2vec} or \textit{GloVe} \citep{Choi2016, Denaxas2018}. 
%These, techniques were further improved by combining dynamic and static data \citep{Choi2016a}, and including the hierarchy of \gls{ehr} data \citep{Choi2017} and the relationships between different medical codes \citep{Choi2018}.
%The drawback of such approaches is that they
%learn an embedding for a specific admission or time window and not for a patient trajectory. 
%This is very challenging in the context of irregular sampled, \gls{ehr} data with a high degree of missingness and multiple data types.
One solution is to use a \gls{rnn}, which has been widely used in the context of supervised prediction tasks such as disease progression \cite{Zhang2018} or more recently in the context of supervised patient stratification \cite{Lee2020}.

% There are also a couple of non-AE based approached using matrix factorization, which could be additionally included 
Few studies have focused on learning a meaningful embedding for unsupervised phenotyping, which can be achieved using an \gls{ae} approach. 
Baytas and colleagues proposed to use a \gls{tlstm} \gls{ae} network \cite{Baytas2017}. 
By explicitly including the time difference between admissions, they showed that they can learn a superior embedding compared to a simple \gls{lstm} using fixed time windows and imputation of missing information.
Yin and colleagues suggested a different approach by using a bi-directional \gls{lstm} with an attention mechanism to focus on the important time windows \cite{Yin2020}. 
This study is one of few which included continuous laboratory values in the analysis, which were imputed if missing or normalized to mean of zero and standard deviation of one. %However, no further details were presented how well binary and continuous information were encoded in the embedded space.
On the contrary the vast majority of publications only consider binary data \cite{Baytas2017, Kim2020, Afshar2019, Beaulieu-Jones2016,Choi2016, Choi2016a}.

To the best of our knowledge, all published unsupervised patient stratification applied to real world \gls{ehr} data focused on the analysis on all available data  \cite{Baytas2017, Yin2020}. Stratification of \textit{sections} of a patient trajectory was not performed. 
However, analysis to answer questions such as \textit{``Are there specific subphenotypes which can be identified using data before the index event, e.g. first occurrence of a heart failure episode? ''} are highly relevant for example, for early stage drug development or to optimize inclusion and exclusion criteria for clinical trials. 

There are a number of papers in the literature that highlight the potential issues of trajectory bias. One paper discusses the general issue of auxiliary objective function deviating from the desired target task during embedding learning and refer to this problem as \gls{ofm} \cite{Metz2019}. Our described problem of trajectory bias is an example of \gls{ofm} where the auxiliary objective function, minimization of the \gls{ae} reconstruction loss, does not match the primary objective, clustering. The information of the trajectory length is essential in achieving a small reconstruction error, but can lead to trajectory length being the dominating factor in the clustering results. While there are no examples in the literature that highlight this problem in \gls{ehr} data, there are examples from other domains that utilise time-series data such as speech recognition. Trosten and colleagues carried out clustering on speech recordings of variable length and found that using sub-optimal embedding representations led to the length of a given speech sequence influencing it's relative position in the embedding space \cite{Trosten2019}. Similarly, Lenco and colleagues demonstrated that data length can dominate clustering methods when used in speech and activity recognition datasets with variable input data length \cite{Lenco2020}.

\section{Methods}

\subsection{Dataset, cohorts and patient trajectories}
\label{sec:datasets}

\textbf{Dataset.} The dataset used in the present study consists of two 
cardiovascular patient cohorts provided by 
% SELECT ON FOR ANONYMISATION
\gls{ouh}
%ANONYMISED
and \gls{chelwest}, to each of which we refer as the \textit{wider cohort}. 
Patients were included in this cohort based on the presence of any cardiovascular specific diagnosis, procedure or medication code throughout their patient history.  
The \gls{ouh} \textit{wider cohort} contains a total of $493,512$ patients, with records spanning from August 2014  to March 2020, and the \gls{chelwest} \textit{wider cohort} a total of of $129,852$ patients, with records spanning from February 2014  to March 2020. 
The \gls{ehr} data contains 4 separate types of clinical features: diagnosis codes (ICD-10 codes), procedure codes (OPCS-4 codes), medication codes (BNF codes), and laboratory values. 
Diagnosis codes could either appear in the data as a primary diagnosis (indicating the primary reason for the hospital admission) or secondary diagnosis (further present comorbidities). 
While the majority of these are binary or categorical features, laboratory values are continuous. 
In addition to this, administrative information, such as start date of admission, discharge date and admission type, were present.

\textbf{Disease specific cohorts and patient trajectories.} To investigate specific patient trajectories, we defined smaller cohorts with an \textit{index event}, as shown in \hyperref[fig:sketch_bias]{Figure 1}.
In this analysis, we defined the cohorts based on two diseases: \textit{acute \gls{hf}} or \textit{stroke}. The smaller cohorts are a subset of the \textit{wider cohort} for each trust, for whom the first instance of the diseases could be detected. This event was our \textit{index event}. The inclusion and exclusion criteria for the first acute \gls{hf} or stroke event are detailed in appendix \ref{apd:hf_event}. 

In principle there can be 5 different trajectories investigated relative to a single \textit{index event}: 
\begin{itemize}
    \item \textit{\Gls{bee}}: the trajectory consists of all data available before the \textit{index event},
    \item \textit{\Gls{s2e}}: the trajectory consists of all data available before and at the \textit{index event} (as sketched in \hyperref[fig:sketch_bias]{Figure 1}),
    \item \textit{\Gls{e2e}}: the trajectory consists of all data available at and after the \textit{index event}, 
    \item \textit{\Gls{afe}}: the trajectory consists of all data available after the \textit{index event}, and
    \item \textit{\acrshort{all}}: considering the complete available data  of the patient trajectory independently of the \textit{index event} (maximum from 2014-2019).
\end{itemize} 
Each of these trajectories can be relevant to answer different clinical questions. 
For instance, phenotyping of \gls{afe} and \gls{e2e} trajectories 
could reveal groups of patients with an unmet medical need indicated by a lower survival rate or be used to model disease progression. 
In contrast, phenotypes based on data of a \gls{bee} or \gls{s2e} trajectory can be used to support clinical trials if e.g. the \textit{index event} aligns with the point of recruitment. 
As this study focuses on the impact of the trajectory bias from a methodological point of view, this paper only includes results of the \gls{e2e}, \gls{afe}, and \acrshort{all} trajectory. 
Similar effects were found using \gls{bee} and \gls{s2e} trajectories.

\hyperref[tab:ouh_stats]{Table II} and \hyperref[tab:chelwest_stats]{Table III} in appendix \ref{apd:cohort_stats_appendix} provide a brief statistical overview of the considered cohorts and trajectories.

%Within the full sample of patients, distinct cohorts can be generated using a fixed set of criteria to represent subgroups of patients with a shared clinical profile. Two clinical cohorts were defined for the current analysis; one that included patients with a history of stroke and one that included patients with a history of heart failure. The full criteria for these two cohorts were defined by a set of clinicians before extracting the relevant patients from the data.

%The inclusion criteria for the stroke cohort were patients over the age of 18 with either an inpatient or A\&E admission for the primary ICD-10 code of I63* Cerebral infarction, who had additional data at least 6 months prior to this event. Additional exclusion criteria were then applied to remove those patients who had primary or secondary ICD-10 codes for either I63* Cerebral infarction or I69.3 Sequelae of cerebral infarction that occurred prior to the first recorded admission for ischaemic stroke, as defined in the inclusion criteria.

\textbf{Data preprocessing.}
There is the risk of implicitly encoding the number windows with data when training the autoencoder using sequences of different lengths. Therefore, we decide to use equal length sequences for all patients. 
Each trajectory of a patient $j$ was divided into non overlapping time windows $x_{j,i}$ of 90 days, with $i$ being the time index. 

As the data spanned more than five years, this resulted in up to $i_{max}= 22$ windows per patient. Features with an occurrence of $<1\,\%$ within the \textit{wider cohort} or the smaller cohorts were removed.

Features were extracted per time window if data was present. Time windows with no data of any type were filled with an \textit{empty} vector consisting in zeros for the binary features, and -0.1 for the normalized continuous features.

The binary features (primary and secondary diagnosis, procedures and medication codes) were included using multi-hot encoding.
Even though continuous laboratory values contain clinically relevant information, the integration of these in an unsupervised \gls{rnn}-\gls{ae} model and combination with further binary data is challenging, as the data need to be normalized and missing values imputed. 
Hence, such data are often not included in large scaled \gls{ehr} data studies.
We explored the effects of applying different data transformations for continuous laboratory values to determine which is most suitable for an \gls{ehr} dataset comprised of mixed data types. 
The approach used was rank normalization \cite{Qiu2013}, where values for a given laboratory measurement were ranked according to all values in the \textit{wider cohort} and then the ranks were normalized to the range $[0,\,1]$. We first applied the rank normalization to the raw data (before doing any time window aggregation). Then, for each individual laboratory type its values within a time window $x_i$ were encoded using 6 features: \textit{min}, \textit{max}, \textit{mean}, \textit{median absolute deviation (MAD)} as well as the last value within the time window and number of occurrences per time window.  If no values were present within a time window for a given laboratory type it was masked as $-0.1$.

After filtering, the total number of different features were partitioned into categories: 393 primary diagnosis codes, 477 secondary diagnosis codes, 202 procedure codes, 126 medication types, and 58 laboratory values for \gls{ouh} and 468 primary diagnosis codes, 586 secondary diagnosis codes, 254 procedure codes, 227 medication types, and 1362 laboratory values for \gls{chelwest}.
Additionally, the number of admissions and the percentage of days in hospital per time window were included.

\begin{figure}[tb]
\label{fig:model} 
 {\caption{Illustration of the \glsfirst{agru} model. A \gls{gru} autoencoder model is extended by a discriminator loss $L_D$ to compensate the impact of the trajectory bias within the patient embedding $Z$. (FC - fully connected layer)}} 
 {\includegraphics[width=1\linewidth]{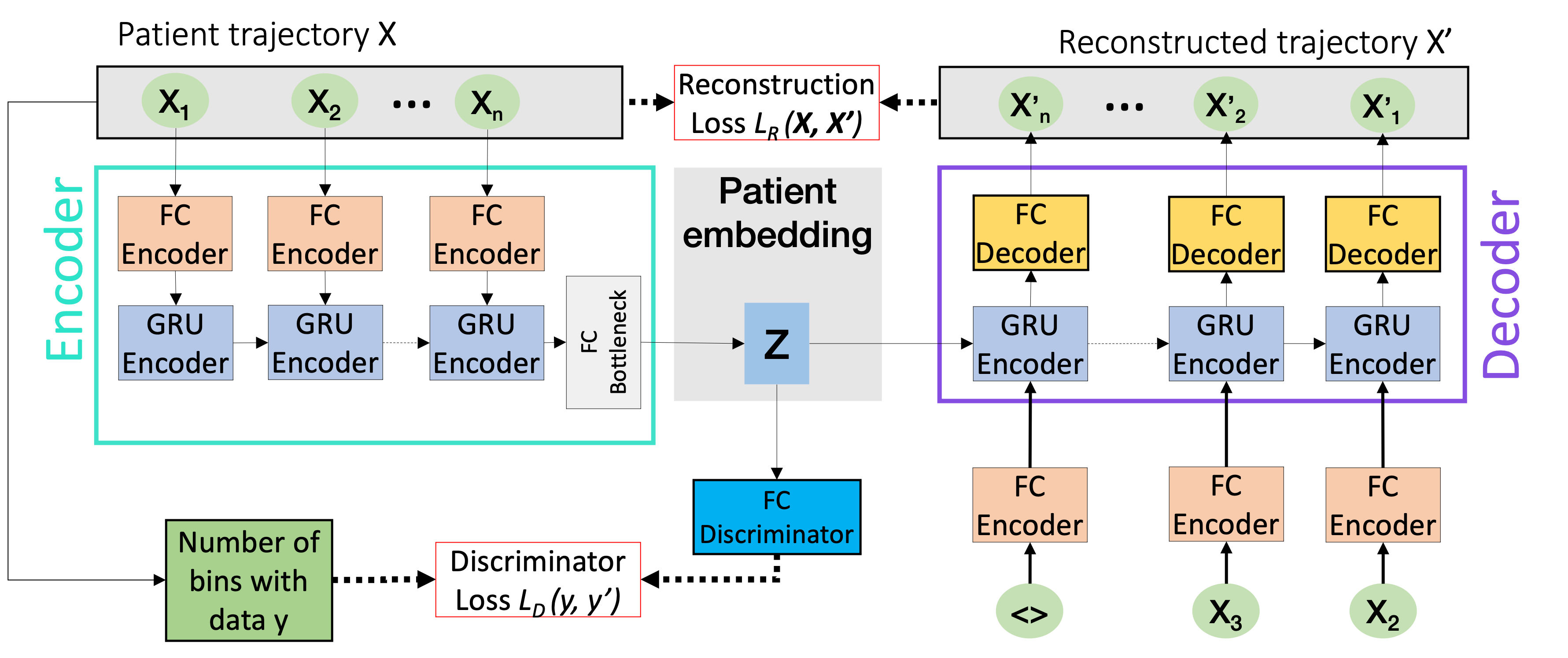}}
\end{figure}

\subsection{Models}
\label{sec:models}
\textbf{Learning a temporal patient embedding.} Our proposed \gls{agru} model is illustrated in \hyperref[fig:model]{Figure 2} and is based on a standard \gls{rnn}-\gls{ae} architecture, which was further extended by an adversarial training scheme as proposed in \cite{Ganin2016}. 
The encoder consists of a feature embedding layer which transforms the features of time window $\textbf{x}_i$ into a fixed-sized embedded vector of $256$ dimensions using a \gls{fc} layer with ReLU activation function.
The embedded time windows are passed into a bidirectional \gls{gru} layer, finally the last hidden state is feed into a bottle neck fully connected layer which transforms the sequential data into a patient embedding $z$ with an  dimension of $n_z\,=\,256$.
The decoder consists of a unidirectional \gls{gru} model followed by a \gls{fc} feature decoder layer which reconstructs the trajectory $\textbf{X}'$.
The \gls{ae} model is learned by minimising the reconstruction loss $L_R$ which is defined as:
\begin{equation}
    L_R(\pmb{X}, \pmb{X'}) = w_b * L_{bin} + L_{cont}
\end{equation}
where $L_{bin}$ and $L_{cont}$ are the loss functions for the binary and continuous features and $w_b$ is a constant. Empirical results indicated that a good reconstruction performance for both, binary and continuous data, can be achieved by weighting the binary loss terms by $w_b=100$. Both losses use \gls{mse} to calculate the error as follows: 

\begin{equation}
    L_{bin/cont} = \frac{1}{N}\sum_{i \in W_{T}}\sum_{j \in F_{bin/cont}} (\pmb{X}(i, j) - \pmb{X'}(i, j))^2
\end{equation}

% \begin{equation}
%     L_{cont} = \frac{1}{N}\sum_{i \in W_{T}}\sum_{j \in F_{cont}} (\pmb{X}(i, j) - \pmb{X'}(i, j))^2
% \end{equation}

where $N$ is the total elements used in the summation, $W_{T}$ refers to the windows inside the sequence that contains data (patient trajectory windows), and $F_{bin/cont}$ are the indices of the binary or continuous features respectively. Additionally, for $L_{cont}$ all the masked missing values where $\pmb{X'}(i, j) < 0$ are excluded as well.
We refer to this model, without the adversarial training scheme, as the \gls{gru} model. 

To compensate the impact of the trajectory bias, this model is extended by an adversarial training scheme.
The patient embedding $z$ is additionally connected to a \gls{fc} discriminator layer which predicts the number of time windows with data  which is optimized by 
\begin{equation}
    L_D = \frac{1}{N}(n_w - n_w')^2
\end{equation}
where $n_w$ is fraction of windows with data within a sequence (trajectory length over sequence length), and $n'_w$ is the discriminator layer prediction.

The autoencoder loss is then modified as
\begin{equation}
    L'_R = L_R - \alpha \, *\, min(\beta, L_D) \label{eq:discriminator}
\end{equation}

with $\alpha$ being a weighting parameter to control the impact of the adversarial training scheme, and $\beta$ is a threshold to cap the influence of $L_D$ once it reaches a certain value. The discriminator ($L_D$) and autoencoder losses ($L'_R$) are minimized sequentially using the Adam optimizer with a learning rate of $2 \times 10^{-3}$ for the autoencoder and $2 \times 10^{-4}$ for the discriminator, and a weight decay of $10^{-6}$ for both of them.

Our adversarial training approach is based on work by Ganin and colleagues \cite{Ganin2016} with the following modifications: 1) we use a regression task instead of a classification one, 2) we optimize the generator and discriminator in two separate steps instead of using a gradient reversal layer, 3) we use a fixed value for $\alpha$ for the whole training, and 4) we cap the discriminator influence as $min(\beta,L_d)$ to stabilize the learning process.

The models were trained using 80\% of the data of the \textit{wider cohort} using the remainder for validation. The trained models were then used to extract the embedding on the specific cohort where the trajectories were aligned on the index event.

To further investigate the impact of the trajectory bias, we also implemented a \gls{tlstm} model as proposed by Yin and colleagues \cite{Yin2020}. For this model we  removed the windows without data from the input, and add as input the time difference between the remaining windows with data.

\textbf{Patient clustering.} In the second step, a clustering of the patients in the embedded space is performed. 
Different clustering algorithms can be used. 
For the purpose of demonstrating the trajectory bias effect on the clustering results we present the results of a simple k-means algorithm. 
Therefore, patient embedding $z$ was reduced to $6$ dimensions using \gls{pca}, as k-means does not perform well in high dimensional spaces \cite{aggarwal2001surprising}.
% Note, that we have also applied this method using different clustering algorithms such as deep embedded clustering \cite{Xie2016}, obtaining similar results with respect to the trajectory bias (results not shown). 
\begin{figure*}[!tb]
\begin{center}
\label{fig:exp2_trajectory_bias}
  {\caption{Visualization of the first two principal components of the embedding space for a a) normal \gls{gru}, b) \gls{tlstm} and c) \glsfirst{agru} \acrlong{ae} model. Each dot represents the embedding of a patient in the $wider$ cohort. The color coding indicates how many time windows of the specific patient trajectory contains any kind of data. The legend shows colors for a selected number of values from the full range of possible trajectory durations.}}
  {\includegraphics[width=0.9\linewidth]{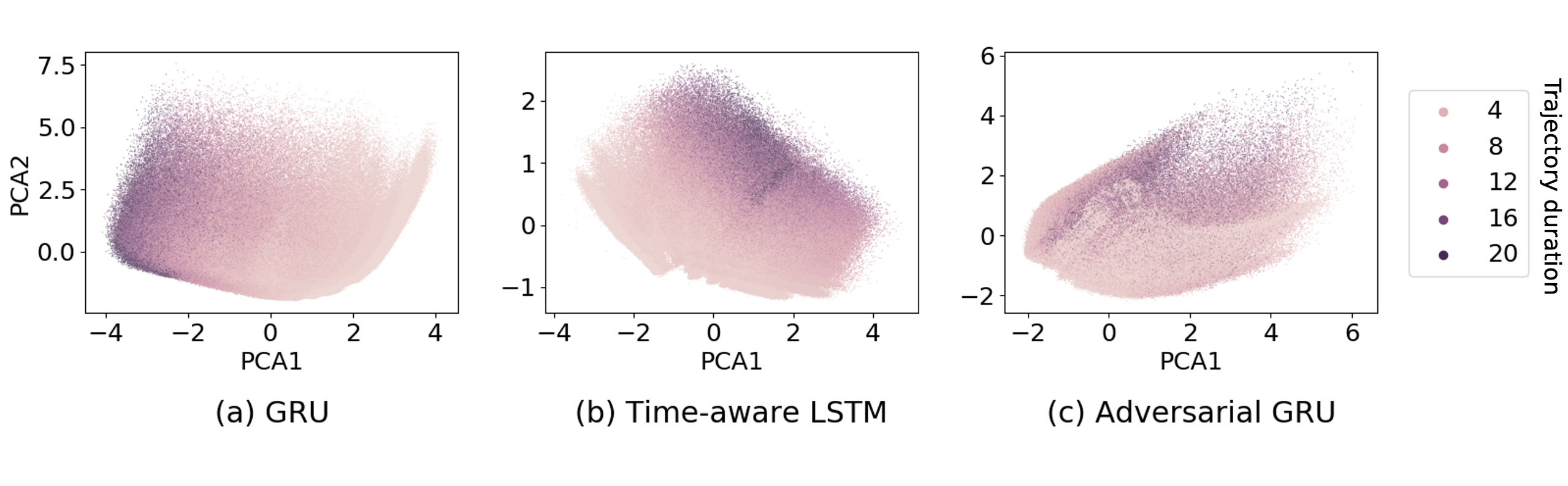}}
\end{center}
\end{figure*}

\subsection{Evaluation}
\label{sec:evaluation}

To quantify how much a given patient embedding is influenced by trajectory bias, we took the \gls{rmse} of the differences between the trajectory length for the patient of interest and the trajectory lengths of the nearest neighbors in the embedding space obtained using a \gls{knn} regression model. We also quantified the extent of trajectory bias in the embeddings by training a surrogate model to predict cluster assignments using only information regarding which time windows were present for patients in the cohort and then evaluating the precision scores. Details of both the \gls{knn} error and the surrogate model method are given in appendix \ref{apd:eval_methods}.  

\section{Results}
\begin{figure*}[!tbp]
\begin{center}
\label{fig:exp3_trajectories}
  {\caption{Comparison of the patient embeddings using the \gls{hf} cohort and the 3 trajectories:  \acrlong{afe}, \acrlong{e2e}, \acrshort{all} for a normal \gls{gru} model and our proposed \gls{agru} approach ($\alpha=1$). Color coding indicates number of time windows per patient with any type of data (trajectory length). The blue box in case of \gls{gru} model indicates a cluster of patients which were only admitted to the hospital for the acute \gls{hf} episode. This group is stronger overlapping with other patients in case of the \gls{agru} model.}}
  {\includegraphics[width=0.7\linewidth]{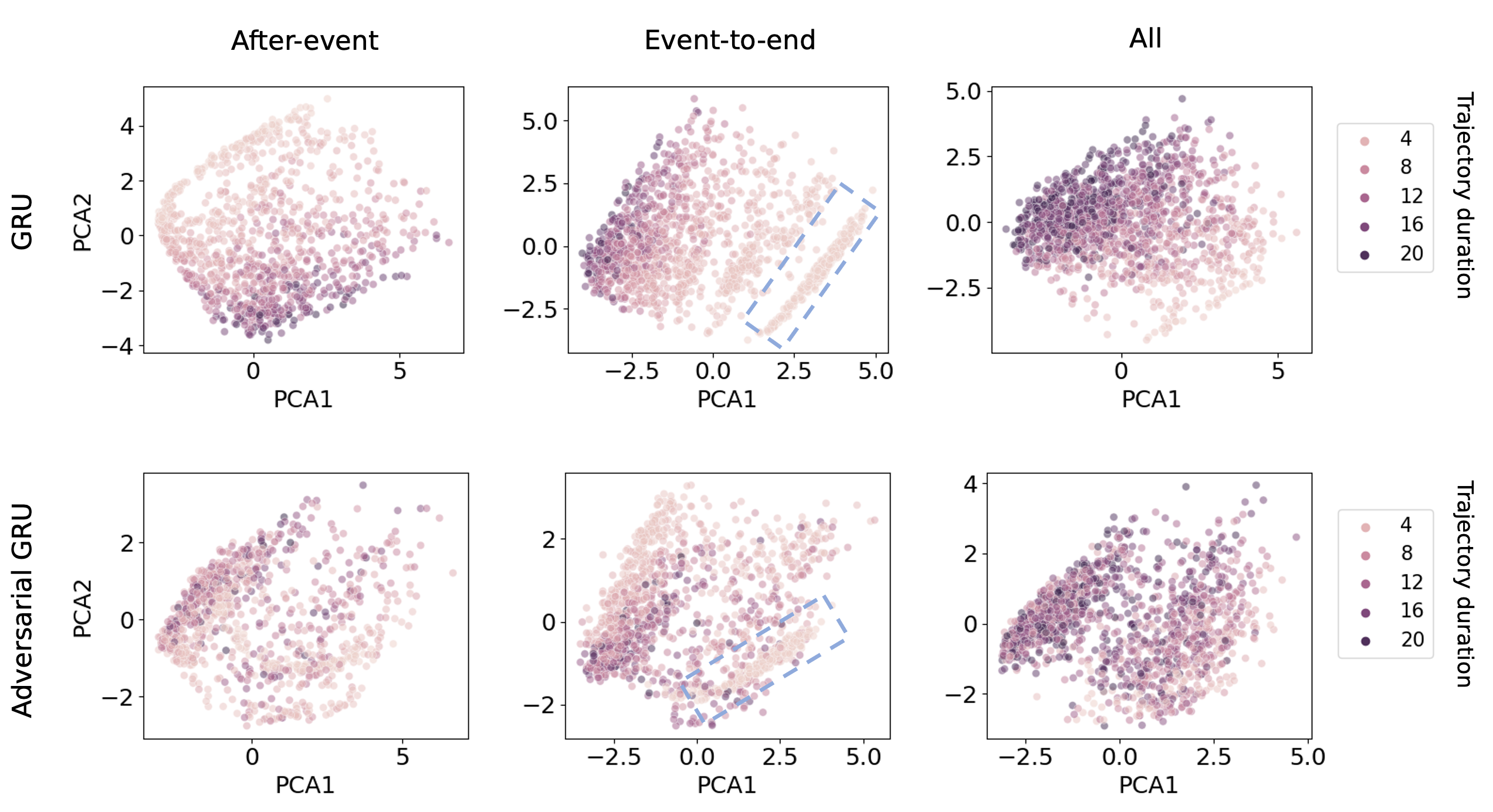}}
\end{center}
\end{figure*}

\textbf{\textit{Wider} cohort evaluation:}
To investigate this effect independently of specific trajectories, we trained our proposed  \gls{agru} model on the $wider$ cohort and compared the results of the patient embeddings with a normal \gls{gru} and a \gls{tlstm} model. 
The models were trained using all 5 data types (primary and secondary diagnosis, procedures, medications and laboratory values). 
The continuous data is included using rank normalization. 
The adversarial training scheme was implemented using $\alpha=1$ and $\beta=0.01$.

\hyperref[fig:exp2_trajectory_bias]{Figure 3} shows a visualization of the first two dimensions of \gls{pca} of the embedding space for all three models. 
Each point represents a patient in the \textit{wider} cohort. 
The color coding indicates the number of windows which contain any type of data per patient (trajectory length). 
The results indicate a strong correlation between trajectory length and the first principal components for the \gls{gru} and a \gls{tlstm} model, which confirms our hypothesis.
Using our adversarial training scheme, the impact on the trajectory length can be reduced.
We quantified our findings using our metric for local differences in trajectory length measured in the embedding space (see \hyperref[sec:evaluation]{section 3.C}). The \gls{knn} error (calculated in the \gls{ouh} wider cohort) was 0.99 (std = 0.05) for the \gls{gru}, 1.46 (std = 0.09) for the \gls{tlstm} and 2.78 (std = 0.12) for the \gls{agru}. The increased value in the \gls{agru} model is indicative of a reduced influence of trajectory length.
%AJ Should we add ChelWest?

% OP TODO: Update values for wider cohort

\textbf{Smaller cohort evaluation:}
We further investigated the trajectory bias in the specific cohorts. For each hospital we evaluate the models trained on the corresponding \textit{wider} cohort on the smaller patient subset defined by {\it\gls{hf}} or {\it stroke} cohort. We aligned the patient trajectories to their index event using three trajectory types:  \gls{afe}, \gls{e2e}, \acrshort{all}.
For the \gls{hf} cohort, the results were generated for different weighting parameters $\alpha$ and clipping parameter $\beta$ to investigate the impact on the embedding results. Additionally, a simple cluster analysis as described in sec \ref{sec:models}.

\hyperref[fig:exp3_trajectories]{Figure 4} shows the first two \gls{pca} axis of the learned embedding space for the different trajectories of the \gls{hf} cohort. 
As in case of the \textit{wider} cohort analysis, the trajectory bias effect is clearly visible in case of the \gls{gru} model. 
The extent to which the embedding model was biased by the input trajectory lengths was again quantified using \gls{knn} error (see \hyperref[sec:evaluation]{section 3.C}). The \gls{knn} error values (calculated in the \gls{ouh} HF cohort) for the \gls{gru} were 1.44 (std = 0.02), 1.23 (std = 0.01) and 2.48 (std = 0.01) for the \gls{afe}, \gls{e2e}, and  \acrshort{all} trajectories respectively. For the \gls{agru} model the \gls{knn} error values were 2.94 (std = 0.03), 2.19 (std = 0.05) and 3.13 (std = 0.01) for the \gls{afe}, \gls{e2e}, and \acrshort{all} trajectories respectively. Again, the increased values for the \gls{agru} model are indicative of a reduced influence of trajectory length. Note that the values obtained in the HF sample were larger than the values obtained in the wider cohort due to the reduced number of patients. Similar results were obtained for the other cohorts analyzed and are summarized in the appendix \ref{apd:results}.

For instance in case of \gls{e2e} trajectory, a clearly separable group of patients with very low number can be identified (marked by the blue box).
These patients have only a single time window with data, indicating that they got only admitted for the acute \gls{hf} event to the hospital and were further treated in a different facility.
The trajectory bias can be compensated using the proposed adversarial training scheme (see lower row of the plots). 
As a result for the \gls{e2e} trajectory embeddings, the before mentioned group of \gls{hf} patients with a single time window is stronger overlapping with patients with longer trajectories. 
\begin{figure*}[!tbp]
\label
  {fig:exp3_a}
  {\caption{Comparison of the embedding space using different weighting factors $\alpha$ for the \acrlong{afe} trajectory of the \gls{ouh} \gls{hf} cohort. Color coding indicates the number of time windows with data per patient.}}
  {\includegraphics[width=1\linewidth]{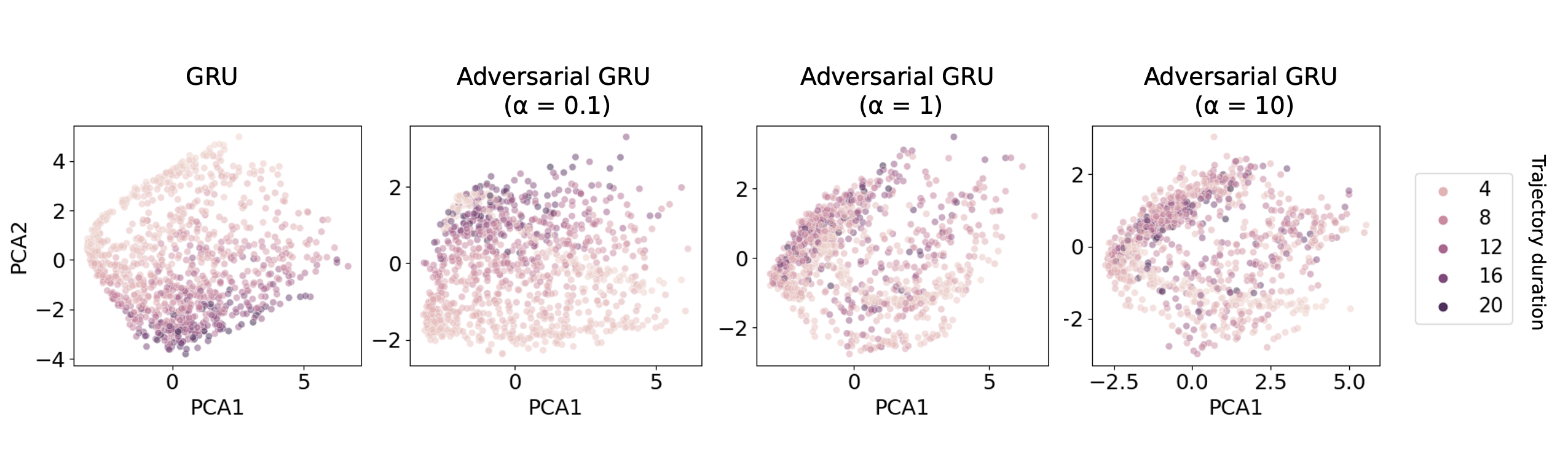}}
\end{figure*}

The impact of the adversarial training can be adjusted by hyperparameter $\alpha$ (see Equation \ref{eq:discriminator}) which serves as a weighting parameter. 
The effect on the patient embedding space is visualized for $\alpha=\{0.1,\, 1, \, 10\}$ and compared to a normal \gls{gru} model for the \gls{afe} trajectory. 
Similar results were found for the other trajectories.
It can be observed from the patient embedding plots in \hyperref[fig:exp3_a]{Figure 5} that the trajectory bias decreases as the value of $\alpha$ increases. This effect can also be seen by looking at the calculated \gls{knn} error values for the different alpha values (see \hyperref[sec:evaluation]{section 3.C}). The \gls{knn} error (calculated for the \gls{afe} trajectory in the HF cohort) for the \gls{gru} was 1.44 (std = 0.02), while the values calculated for the \gls{agru} models were 2.69 (std = 0.02), 3.14 (std = 0.04) and 3.18 (std = 0.02) for $\alpha = 0.1$, $\alpha = 1$ and $\alpha = 10$ respectively. 

Initial results using the adversarial training scheme as proposed by Ganin and colleagues \cite{Ganin2016}, resulted in an unstable training due to high gradients caused by $L_{adv}$. To compensate this effect, we constrained the adversarial training impact using $\beta$ (see Equation \ref{eq:discriminator}. 
Our results indicate that a learning is possible using a $\beta = \{0.1, \,0.01\}$. 
The parameter $\beta$ was set to $0.01$ for all results using \gls{agru}.
An alternative solution to this problem is the clipping of gradients. 

How strong the trajectory bias should be compensated strongly depends on the specific clinical question and dataset. One way of determining how the $\alpha$ and $\beta$ values should be tuned is through evaluation of performance on a relevant downstream task.

The influence of the trajectory bias on potential patient clusters was further evaluated using a k-means cluster algorithm. 
As an example, the clustering was performed with $k=6$ clusters.
The cluster results, visualized in the embedding space, for the \gls{ouh} \gls{hf} cohort are shown in top row of \hyperref[fig:exp3_clustering]{Figure 6}. 
The box plots below show the distribution of the number of time windows per cluster.
In case of the clustering results using the \gls{gru} model, it can be observed that the cluster $3$ only focuses on patients with a very short trajectory and in contrary cluster $2$ and $4$ on patient with a long trajectory. 
The difference between the clusters is strongly compensated using the \gls{agru} approach.
%Similar finding were also achieved using other clustering algorithms such as deep embedded clustering \citep{Xie2016}.

A further quantification of the adversarial training scheme was performed by using surrogate prediction approach as described in \hyperref[sec:evaluation]{section 3.C}.
\hyperref[tab:exp3_surrogate]{Table I} shows the averaged precision values for predicting the cluster assignment given the number of time windows with data for different values of $\alpha$ and trajectories. 
The average precision scores are higher for the \gls{e2e} trajectory due to presence of patients with only a single time window with data, as discussed in the context of \hyperref[fig:exp3_trajectories]{Figure 4}.

\section{Discussion}

The results presented within the previous section indicate the impact of the trajectory bias for different methods, cohorts and trajectories. 
As shown in \hyperref[fig:exp3_trajectories]{Figure 4}, a typical \gls{gru} implementation as well as a \gls{tlstm} approach are effected.
In case of the latter, the time difference between time windows is directly included in the \gls{rnnae} model.
For instance, if a patient has only data in the first quarter of 2016 and in the third quarter of 2019 the input sequence will have two elements, empty time windows in-between will be ignored. By contrast the \gls{gru} model input will have a fix sequence size of 22 where all the empty time windows are filled with the \textit{empty} admission vector.
This results in an decreased trajectory bias (indicated by the higher \gls{knn} error value of $1.46$ compared to $1.08$ for the \gls{gru} model).
Nonetheless, the trajectory bias remains present as patients with more data will result in a longer input sequence. 
This becomes visible in \hyperref[fig:exp2_trajectory_bias]{Figure 3}.b as the first two principle components of the patient embeddings $z$ are highly correlated to the trajectory length (see \hyperref[fig:model]{Figure 2}). 
Our proposed adversarial training scheme approach can compensate this effect. 
This is visually confirmed in \hyperref[fig:exp2_trajectory_bias]{Figure 3}.c and \hyperref[fig:exp3_trajectories]{Figure 4} (which are limited to only the first two principle components of the embedding space) as well as quantitatively with the \gls{knn} error and surrogate model (see \hyperref[tab:knn_error_all]{Table V} in appendix \ref{apd:results}).
Please note, the adversarial training scheme is not limited to a \gls{gru} model and could also be applied to the \gls{tlstm} approach.  
\begin{figure}[tbp]
\label
  {fig:exp3_clustering}
  {\caption{Top row: k-means clustering results on the patient embedding and bottom row: box plot of distribution of number of windows with data per cluster using either a a) \gls{gru} or b) \gls{agru} \acrlong{ae} model.}}
  {\includegraphics[width=1\linewidth]{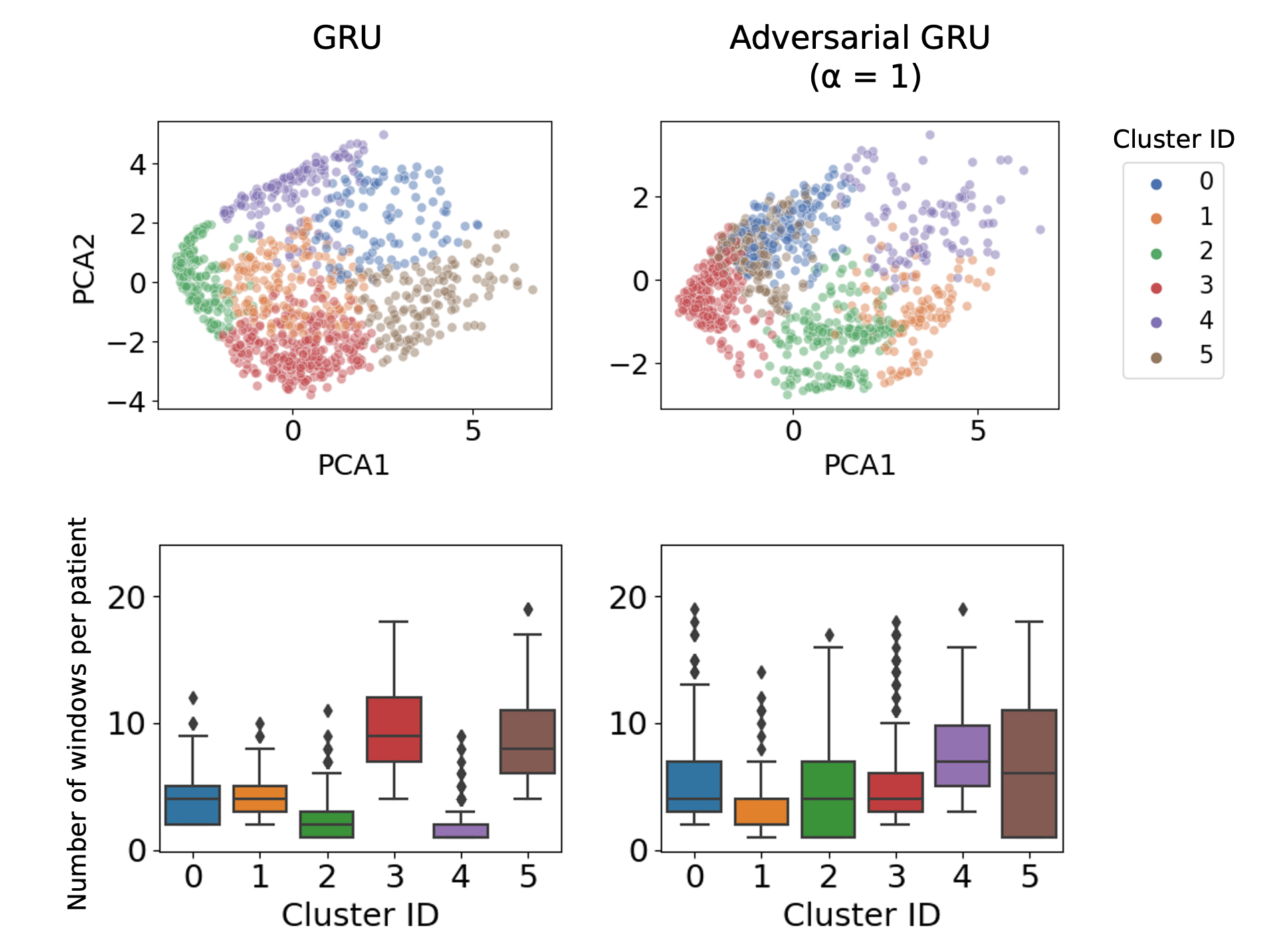}}
\end{figure}
\begin{table}[tb]
\resizebox{\columnwidth}{!}{
\begin{tabular}{lccc}
  \toprule
\bfseries Model   & \bfseries \glsfirst{afe}    & \bfseries \glsfirst{e2e} & \bfseries All  \\
  \midrule
\gls{gru}            & 0.44±0.03 & 0.64±0.03 & 0.49±0.04 \\
\gls{agru} ($\alpha=0.1$)      & 0.38±0.04 & 0.57±0.02 & 0.41±0.03 \\
\gls{agru} ($\alpha=1$)        & 0.34±0.05 & 0.55±0.03 & 0.40±0.05 \\
\gls{agru} ($\alpha=10$)       & 0.30±0.04 & 0.37±0.02 & 0.40±0.04 \\
  \bottomrule
\end{tabular}
  }
{\caption{\label{tab:exp3_surrogate} Average precision scores of a surrogate model to predict the cluster assignment (derived using k-means clustering) on the OUH HF cohort  for different trajectories: \acrshort{afe}, \acrshort{e2e}, and \acrshort{all} for different \acrshort{ae} models. Lower values here indicate that the embedding model has a reduced trajectory bias.}}
\end{table}

Our objective is that by compensation of the trajectory bias, an unsupervized phenotyping approach is more likely to find patient clusters which are determined by specific clinical observations rather when only the trajectory length.
However, we emphasize that the knowledge of how much data is available in a patient trajectory is an important information in general.
A healthy patient will be admitted less frequent to the hospital and vice versa. 
Nonetheless, there are various non-health related reasons, why a patient has less data. This could include reasons such as the patient moved to a different place, different ways of working between wards or modification of the hospital IT infrastructure. As these reasons are often unknown, we want to limit the impact of the trajectory length on the patient embedding.

How strong the trajectory bias should be compensated depends on the specific application. 
We explored the parameter $\alpha$ which serves as a weighting parameter to compensate the trajectory bias (see \hyperref[fig:exp3_clustering]{Figure 6} and \hyperref[tab:exp3_surrogate]{Table I}). 
It needs to be further investigated if there are some generic measures to supports the optimization of $\alpha$ to reduce the burden of clinicians to evaluate large number of cluster results.  

Another aspect, which was not further explored into detail is the \textit{density of data points} within a time window. 
The patient trajectory is transformed into an input sequence consisting of non overlapping time windows of 3 months. 
More granular information is only provided to the model for continuous data (number of tests in a given time window is computed) and in form of two administrative features,  indicating the number of admissions and percentage of days in hospital. 
However in case of binary data, the model does not know if e.g. only one or multiple MRI scans (procedure code) were performed. 
This is of course also an indication of the severity of medical condition, which should be explored further and its relationship to the trajectory bias. 

\section{Conclusion}

We showed that patient embeddings using previously proposed \gls{rnnae} models are impacted by a \textit{trajectory bias}, meaning that 
embeddings are heavily dependent on the amount of data contained in each patients trajectory, likely at the expense of clinically relevant details.
%In particular, the variation in the amount of data recorded across patients can lead to the embedding models being strongly dominated by the amount of data contained in each patients trajectory, instead of encoding clinical relevant details.
We presented a novel method to overcome unwanted \textit{trajectory bias} using an adversarial training scheme on top of a \gls{rnnae}. %that result in a reduced \textit{trajectory bias} in the learnt embedding space. 
We demonstrated that our approach was effective in different datasets, cohorts and a range of different patient trajectory types, with our results showing that we were able to flexibly reduce the extent of \textit{trajectory bias} in our model. 

\acks{
This work uses data provided by patients collected by Chelsea and Westminster Hospital NHS Foundation Trust and  Oxford University Hospitals NHS Foundation Trust as part of their care and support. We believe using the patient data is vital to improve health and care for everyone and would, thus, like to thank all those involved for their contribution. The data were extracted, anonymized, and supplied by the Trust in accordance with internal information governance review, NHS Trust information governance approval, and the General Data Protection Regulation (GDPR) procedures outlined under the Strategic Research Agreement (SRA) and relative Data Processing Agreements (DPAs) signed by the Trust and Sensyne Health plc.
 
This research has been conducted using the Oxford University Hospitals NHS Foundation Trust Clinical Data Warehouse, which is supported by the NIHR Oxford Biomedical Research Centre and Oxford University Hospitals NHS Foundation Trust. Special thanks to Kerrie Woods, Kinga Várnai, Oliver Freeman, Hizni Salih, Steve Harris and Professor Jim Davies.
}

\bibliographystyle{IEEEtran}
\bibliography{references}

\appendices

\section{Statistical description of cohorts and trajectories}
\label{apd:cohort_stats_appendix}
An overview of the considered cohorts and trajectories for are shown in \hyperref[tab:ouh_stats]{Table II} for \gls{ouh} and \hyperref[tab:chelwest_stats]{Table III} for \gls{chelwest}. The number of patients are reported for the wider cohort, as well as the three different trajectories for both the  \gls{hf} and stroke cohort. For each of the feature types in the dataset, the mean number of that feature present per patient are also reported. Finally, a number of summary statisticals are reported for the trajectory durations.

\begin{table*}[!t]
\small
\label{tab:ouh_stats}
{\caption{  Statistical description of used cohorts and trajectories for OUH. \# refers to the number of patients or features for the different data types present in the \gls{ehr} data.}}
  
\resizebox{\textwidth}{!}{\begin{tabular}{lc|ccc|ccc}
\toprule
& $wider$    
    &  \multicolumn{3}{c}{\gls{hf} cohort}  &  \multicolumn{3}{c}{Stroke cohort} \\
& cohort & \gls{e2e} & \gls{afe} & All & \gls{e2e} & \gls{afe} & All \\
\midrule
Number of patients           & 493512 & 1430 & 1097 & 1430 & 1480 & 1035 & 1480 \\
\midrule
\multicolumn{2}{l}{Data per patient Mean (Median):}  &  &  &  \\
\quad Primary Diag.     & 0.96 (0) & 3.77 (3) & 2.72 (1) & 5.93 (5) & 3.01 (2) & 2.21 (1) & 5.15 (4)\\ 
 \quad Secondary Diag.        & 4.38 (1) & 26.51 (18) & 19.61 (10) & 37.5 (27) & 21.79 (15) & 14.97 (7) & 32.41 (23)\\ 
 \quad Procedures        & 3.05 (1) & 7.10 (5) & 5.65 (3) & 12.48 (10) & 7.88 (6) & 4.37 (2) & 13.15 (11)\\ 
 \quad Medications        & 8.38 (2) & 33.42 (22) & 23.91 (12) & 49.60 (37) & 28.83 (20) & 18.59 (5) & 44.99 (32)\\ 
 \quad Laboratory Values        & 71.55 (43) & 91.81 (67.5) & 88.76 (67) & 202.28 (192) & 78.44 (56) & 78.90 (60) & 182.93 (170) \\ 
\midrule
\midrule
\multicolumn{2}{l}{Time windows with data:}  &  &  &  \\
\quad Mean        & 5.16 & 5.15 & 5.41 & 12.46 & 4.32 & 4.74 &  11.29 \\ 
\quad Median        & 4 & 4 & 4 & 12 & 3 & 4 & 11 \\ 
\quad Min        & 1 & 1 & 1 & 1 & 1 & 1 & 1 \\ 
\quad Max        & 22 & 20 & 19 & 23 & 19 & 18 & 22 \\ 
\bottomrule
\end{tabular}}
\end{table*}

\begin{table*}[!t]
\small
\label{tab:chelwest_stats}
{\caption{Statistical description of used cohorts and trajectories ChelWest. \# refers to the number of patients or features for the different data types present in the \gls{ehr} data.}}

\resizebox{\textwidth}{!}{
\begin{tabular}{lc|ccc|ccc}
\toprule
& $wider$    
    &  \multicolumn{3}{c}{\gls{hf} cohort}  &  \multicolumn{3}{c}{Stroke cohort} \\
& cohort & \gls{e2e} & \gls{afe} & All & \gls{e2e} &\gls{afe} & All \\
\midrule
Number of patients & 129852 & 818 & 566 & 818 & 668 & 429 & 668 \\
\midrule
\multicolumn{2}{l}{Data per patient Mean (Median):} &  &  &  \\
\quad Primary Diag.  & 1.72 (1) & 3.84 (3) & 2.93 (2) & 6.08 (5) & 2.79 (2) & 2.17 (1) & 5.15 (4)\\ 
\quad Secondary Diag.  & 10.48 (5) & 34.69 (24) & 30.07 (20) & 48.51 (38) & 29.28 (19) & 24.39 (15) & 44.66 (32)\\ 
\quad Procedures  & 7.78 (5) & 8.68 (6) & 7.31 (5) & 16.22 (13) & 7.72 (6) & 5.70 (3) & 16.31 (12) \\ 
\quad Medications & 6.45 (2) & 21.15 (14) & 17.00 (9) & 29.68 (21) & 19.48 (13) & 14.21 (7) & 29.42 (21) \\ 
\quad Laboratory Values & 86.62 (52) & 110.53 (76) & 105.26 (75) & 211.67 (174) & 83.97 (60) & 73.97 (50) & 186.92 (142) \\ 
\midrule
\midrule
\multicolumn{2}{l}{Time windows with data:}  &  &  &  \\
Mean & 4.84 & 4.19 & 4.61 & 9.08 & 3.25 & 3.51 & 8.18 \\ 
Median & 3 & 3 & 4 & 8 & 2 & 3 & 7 \\ 
Min & 1 & 1 & 1 & 1 & 1 & 1 & 1 \\ 
Max & 28 & 22 & 21 & 27 & 19 & 18 & 27 \\ 
\bottomrule
\end{tabular}}
\end{table*}

\section{Inclusion and exclusion criteria for patients with a first acute heart-failure event}
\label{apd:hf_event}
This event was defined as the occurrence of any of the following ICD-10 codes as a primary diagnosis: (i) I50* Heart failure, (ii) I11.0 Hypertensive heart disease with (congestive) heart failure, (iii) I13.0 Hypertensive heart and renal disease with (congestive) heart failure or (iv) I13.2 Hypertensive heart and renal disease with both (congestive) heart failure and renal failure. In addition to having one of these diagnoses, patients were only included in the cohort if they had at least 3 months of data available prior to the first admission where one of these diagnosis codes were recorded. Finally, we excluded patients from the cohort based on the following criteria: (i) patients whose first admission (as defined above) was under 48 hours and had a heart failure related procedure code in the 30 days following the first admission (OPCS-4 codes: K59*, K60*, K61*, K72*, K73*, K74*), (ii) patients who had heart failure related ICD-10 codes recorded as a secondary diagnoses prior to their first admission (ICD-10 codes: I50*, I11.0, I13.0, I13.2), (iii) patients who had been prescribed Eplerenone, Sacubitril with Valsartan or Spironolactone at a dose of either 25mg or 50mg, (iv) patients with a recorded New York Heart Association classification or (v) patients with a recorded ejection fraction under 40\%.

\section{Details for evaluation methods}
\label{apd:eval_methods}
\textbf{K-nearest neighbors error:} To quantify the extent to which a learned patient embedding is locally influenced by the trajectory bias,
%given embedding model was influenced by the length of the input trajectories, 
we calculated the \gls{rmse} between the trajectory lengths and the prediction of a \gls{knn} regression model for a sample of patients. This  gives a measure of the amount of variation there is in trajectory lengths between patients that are close to each other within the embedding space. 
For a given embedding space we will refer to this metric as the \gls{knn} error. 
The full process consisted of the following steps: first we sampled $N$ random patients from a given cohort, then for a given patient, $i$, we find its $K$ nearest neighbors based on their Euclidean distance within the embedding space. The mean trajectory length for the nearest neighbors, $T_{kNN}$, was then calculated as,
\begin{equation}
    T_{kNN_i} = \frac{1}{K}\sum\limits_{k=1}^{K}{T_{k,i}} 
\end{equation}
where  $T_{k,i}$ refers to the trajectory length (time windows with any type of data a long the sequence) for the $k_th$ nearest neighbor of patient $i$.
%We then took the difference between the trajectory length of the reference trajectory, $L_i$, and the mean trajectory length of the nearest neighbors, $L_{kNN}$. Using these values for the $N$ sampled patient trajectories, we can calculate the D-score:
The final \gls{knn} error is calculated by taking the root of the mean of the squared differences between the \gls{knn} predictions and the length for all the patients in the sample as
\begin{equation}
   kNN_{error} = \sqrt{\frac{1}{N}\sum\limits_{i=1}^{N}{(T_{kNN_i} - T_i)^2}}
\end{equation}

For the results presented here, we used $N = 1000$ and $K = 5$. Increased influence of trajectory length in the embedding model will result in lower \gls{knn} error values. To allow for us to obtain the standard deviation of this measure, each time the \gls{knn} error was reported we calculated the value 10 times (obtaining new samples each time) and reported the mean and standard deviation across these 10 values.

\textbf{Surrogate models:} We used a further metric to investigate the impact of the trajectory bias on identified patient clusters.
%the extent to which trajectory bias was present in the embedding space by training surrogate models to use temporal information only from patient trajectories to classify cluster labels that were assigned in the different embedding spaces. 
Patient trajectory data were converted to simple binary vectors that indicated whether or not a patient had any data for a specific time window.  
%and did not hold any information pertaining to any feature codes that were present during those visit. 
These binary vectors were used as inputs for a simple random forest classifier which was trained to predict the cluster labels assigned to patients by one of the investigated methods. %using k-means clustering on the full patient trajectory embeddings (with all visit information embedded). 
%This process was repeated for the \gls{gru} model as well as the \gls{agru} model for three different alpha values ($\alpha = 0.1$, $\alpha = 1$ and $\alpha = 10$). 
The average precision scores were used to determine how well the surrogate model could predict the cluster labels using only information of the trajectory length. 
A lower trajectory bias in the identified patient clusters is indicated by a low average precision score.
%For each of the embedding models, an average precision score was obtained for each of the 3 different trajectories: \gls{afe}, \gls{e2e} and \acrshort{all}.

\section{Summary of the adversarial approach for all the different datasets.}
\label{apd:results}

\begin{table}[tb]
\resizebox{\columnwidth}{!}{
\begin{tabular}{lll|ccc}
  \toprule
\bfseries Trust & \bfseries Disease & \bfseries Model   & \bfseries \glsfirst{afe}    & \bfseries \glsfirst{e2e} & \bfseries All  \\

\midrule
Hospital 1 & \gls{hf} & \gls{gru} &  0.45±0.03 & 0.63±0.03 & 0.49±0.06 \\
& & \gls{agru}  &  0.36±0.05 & 0.54±0.02 & 0.41±0.04 \\
\midrule
Hospital 1 & Stroke & \gls{gru} &  0.42±0.02 & 0.70±0.03 & 0.54±0.01 \\
& & \gls{agru} &  0.37±0.03  & 0.62±0.02 & 0.49±0.04 \\

\midrule
Hospital 2 & \gls{hf} & \gls{gru} & 0.38±0.05 & 0.60±0.03 & 0.39±0.04 \\
& & \gls{agru}  & 0.29±0.04 & 0.41±0.03 & 0.34±0.03 \\
  \midrule
Hospital 2 & Stroke & \gls{gru} & 0.38±0.05 & 0.47±0.04 & 0.41±0.04 \\
& & \gls{agru} & 0.24±0.06 & 0.40±0.03 & 0.36±0.02 \\

  \bottomrule
\end{tabular}
  }
{\caption{\label{tab:surrogate_all} Average precision scores of a surrogate model to predict the cluster assignment (derived using k-means clustering) for different trajectories: \acrshort{afe}, \acrshort{e2e}, and \acrshort{all} for different \acrshort{ae} models. Lower values here indicate that the embedding model has a reduced trajectory bias.}}
\end{table}

\begin{table}[tb]

\resizebox{\columnwidth}{!}{
\begin{tabular}{lll|ccc}
  \toprule
\bfseries Trust & \bfseries Disease & \bfseries Model   & \bfseries \glsfirst{afe}    & \bfseries \glsfirst{e2e} & \bfseries All  \\

\midrule
Hospital 1 & \gls{hf} & \gls{gru} &  1.44±0.02 & 1.23±0.01 & 2.48±0.02 \\
& & \gls{agru}  &  2.94±0.04 & 2.19±0.05 & 3.13±0.01 \\
\midrule
Hospital 1 & Stroke & \gls{gru} &  1.23±0.03 & 0.93±0.02 & 2.31±0.02 \\
& & \gls{agru} &  2.92±0.05  & 1.71±0.04 & 2.95±0.03 \\

\midrule
Hospital 2 & \gls{hf} & \gls{gru} & 1.35±0.06 & 1.24±0.03 & 2.53±0.04 \\
& & \gls{agru}  & 2.14±0.07 & 1.91±0.05 & 3.31±0.02 \\

\midrule
Hospital 2 & Stroke & \gls{gru} & 1.14±0.04 & 0.99±0.05 & 2.36±0.08 \\
& & \gls{agru} & 1.88±0.04 & 1.63±0.07 & 3.43±0.04 \\

  \bottomrule
\end{tabular}
  }
{\caption{\label{tab:knn_error_all} Average k-Nearest neighbor error values for different trajectories: \acrshort{afe}, \acrshort{e2e}, and \acrshort{all} for different \acrshort{ae} models. Higher values indicate that the embedding model has a reduced trajectory bias.}}
\end{table}

\end{document}